\documentclass{article}
\usepackage[accepted]{icml2020}

\usepackage{comment}
\usepackage[utf8]{inputenc} % allow utf-8 input
\usepackage[T1]{fontenc}    % use 8-bit T1 fonts
\usepackage{hyperref}       % hyperlinks
\usepackage{url}            % simple URL typesetting
\usepackage{booktabs}       % professional-quality tables
\usepackage{amsfonts}       % blackboard math symbols
\usepackage{nicefrac}       % compact symbols for 1/2, etc.
\usepackage{algpseudocode}
\usepackage{algorithm,amssymb,amsthm}
\usepackage{graphicx}
\usepackage{amsmath}
\usepackage{multirow}
\algdef{SE}[DOWHILE]{Do}{doWhile}{\algorithmicdo}[1]{\algorithmicwhile\ #1} %
\newtheorem{theorem}{Theorem}[section]
\newtheorem{lemma}[theorem]{Lemma}

\newtheorem{corollary}[theorem]{Corollary}
\newcommand{\eat}[1]{}
 {
      \theoremstyle{plain}
      \newtheorem{assumption}{Condition}
  }
\numberwithin{equation}{section}

\DeclareMathOperator*{\argmin}{argmin}
\def \bp {{\mathbf{p}}}
\def \bx {{\mathbf{x}}}
\def \bz {{\mathbf{z}}}
\def \bw {{\mathbf{w}}}
\def \Xl {{X_{\lambda}}}
\def \Xlo {{X_{\lambda_1}}}
\def \Xlt {{X_{\lambda_2}}}
\def \Xld {{X_{\tilde{\lambda}}}}
\def \XT {{X_{T}}}
\icmltitlerunning{Optimal package type selection for e-commerce shipments}
\begin{document}

\twocolumn[
\icmltitle{Think out of the package: Recommending package types for e-commerce shipments}
\icmlsetsymbol{equal}{*}

\begin{icmlauthorlist}
\icmlauthor{Karthik S. Gurumoorthy}{amazon}
\icmlauthor{Subhajit Sanyal}{amazon} 
\icmlauthor{Vineet Chaoji}{amazon}
\end{icmlauthorlist}

\icmlaffiliation{amazon}{India Machine Learning, Amazon, Bangalore, India}
\icmlcorrespondingauthor{Karthik Gurumoorthy}{gurumoor@amazon.com}
\icmlcorrespondingauthor{Subhajit Sanyal}{subhajs@amazon.com}
\icmlcorrespondingauthor{Vineet Chaoji}{vchaoji@amazon.com}

\icmlkeywords{Package type selection, Ordinal enforcement, Discrete optimization, Constrained-Unconstrained formulation equivalence, Trade off parameter selection}

\vskip 0.3in
]
\printAffiliationsAndNotice{}
\begin{abstract}
Multiple product attributes like dimensions, weight, fragility, liquid content etc. determine the package type used by e-commerce companies to ship products. Sub-optimal package types lead to damaged shipments, incurring huge damage related costs and adversely impacting the company's reputation for safe delivery. Items can be shipped in more protective packages to reduce damage costs, however this increases the shipment costs due to expensive packaging and higher transportation costs. In this work, we propose a multi-stage approach that trades-off between shipment and damage costs for each product, and accurately assigns the optimal package type using a scalable, computationally efficient linear time algorithm. A simple binary search algorithm is presented to find the hyper-parameter that balances between the shipment and damage costs. Our approach when applied to choosing package type for Amazon shipments, leads to significant cost savings of tens of millions of dollars in emerging marketplaces, by decreasing both the overall shipment cost and the number of in-transit damages. Our algorithm is live and deployed in the production system where, package types for more than $130,000$ products have been modified based on the model's recommendation, realizing a reduction in damage rate of $24\%$.  
\end{abstract}

\section{Introduction}
\label{sec:introduction}
\begin{figure}
\begin{center}
\includegraphics[width= 0.8\linewidth]{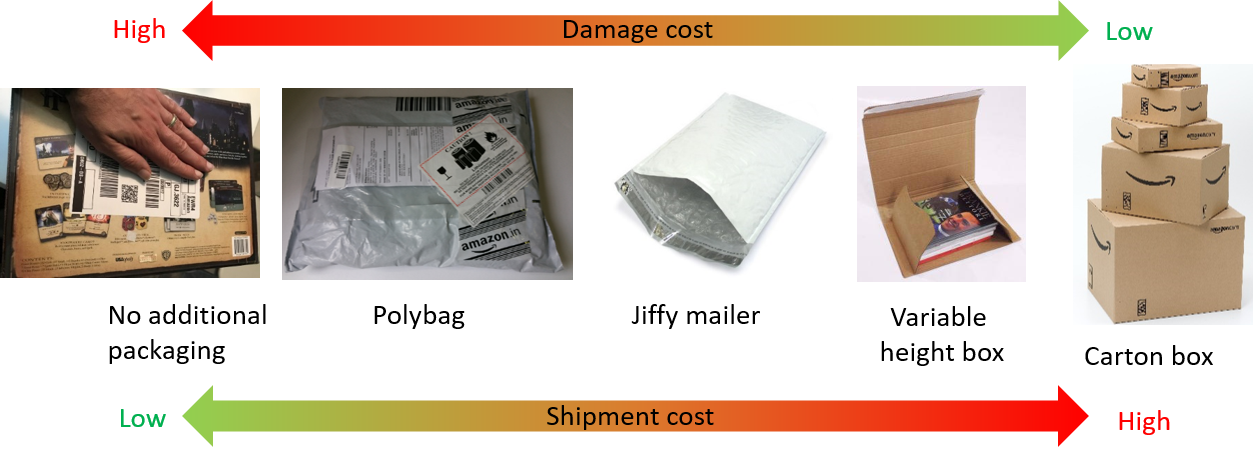}
\end{center}
\caption{Different package types}
\label{fig:pkgtypes}
\end{figure}

E-commerce companies like Amazon uses several different package types to ship products from warehouses to the customer's doorstep. These package types vary in the extent of protection offered to the product during transit. Generally, robust package types that provide more protection to the product --- resulting in reduced number of package related damages --- cost more at the time of shipping due to high material and transportation costs, and vice versa. For instance as shown in Fig.~\ref{fig:pkgtypes}, Amazon has the following different package type options listed in increasing order of protection afforded to the product: (i) No Additional Packaging (NAP), (ii) Polybags: polythene bags small (PS) and special (PL), (iii) Jiffy mailer (JM), (iv) Custom pack (CP), (v) Corrugated T-folder box (T), (vi) Corrugated box with variable height (V), (vii) Corrugated carton box (C). Each package type comes in multiple sizes like small, medium, large and extra-large. The combination of packaging type and size is assigned a barcode, e.g. PS6 to PS9 for small polybags. When an item is ready to be packed, the chosen packaging material and size are used to ship the product.

Damages attributed to packaging can happen during transit or during handling by associates during the shipment from the warehouse to the customer. As damages result in degraded customer experience, an extra amount is often paid to the customer as compensation over and above the product's price. As these damaged products need to be sent back to the warehouse, there is an additional return shipment cost. Such damages adversely affect the customer relationship since the company's reputation for reliable delivery is impacted. For instance, customers who are dissatisfied for time-critical purchases such as during festivities, may hesitate to buy products in the future. There is also an effect on the company's relationship with sellers or vendors or brands, particularly if new product suffers repeated damages over multiple shipments, since the first few customer experiences for such new products are very critical to the seller or vendor for the long term success of the product. The sum of all these costs associated with damages will henceforth be referred to as damage cost.

To reduce the cost of damages, items can be packed in more protective packaging. However, more protective packaging (e.g. corrugated box (C)) costs more in terms of packaging materials and transportation costs, which increase the shipping cost that customers have to pay, or costs that the company bears in case of free shipping. It also generates packaging waste at the customer's end which needs to be disposed off additionally. Hence the important problem that needs to be addressed is: \emph{"What is the right package type say, between polybay, jiffy mailer or different corrugated boxes that should be used for shipping a product with the best trade off between shipping cost and damage cost ?"} Once the package type is chosen, the smallest container (size) of that package type that could fit the product snugly would be used for actual shipment. This reduces the shipment volume and hence the shipping cost. 

\subsection{Contributions}
Below, we list our main contributions in this work:\\
(i) We propose a two-stage approach to recommend the correct package type for products resulting in significant savings, primarily from decreased packaging related damages. Our model's recommendation also leads to decreased shipping cost compared to the current selection of package types, the reason for which is explained in Section~\ref{sec:pkgtyperecommendation}. \\
(ii) Our framework provides a scalable mechanism for the package type recommendation, circumventing manual intervention at every stage and deprecating the existing keyword based approach of mapping package type explained in Section~\ref{sec:currentapproach}, which is slow, reactive and often subjective.\\
(iii) We establish novel theoretical connections between the constrained (Ivanov) and unconstrained (Tikhonov) formulations for our unique setting where the optimization variable is discrete, and show that while the constrained formulation is $NP-$ complete, the unconstrained formulation enjoys a linear time solution. Though such connections based on the Lagrange dual formulation are known when the optimization variable is continuous~\cite{Oneto16}, the proof methodology employed in our work to derive similar equivalences when the optimization variable is discrete requires fundamentally new insights into the solution space. To the best of our knowledge, this connection is unknown and not exploited before.\\
(iv) Our understanding of the solution space further enables us to consistently choose the hyper-parameter for the unconstrained formulation using a simple binary search type algorithm, which optimally provides the trade-off between the different cost parameters. \\
To summarize, we provide a scalable approach for choosing the best package type for products and also present an efficient algorithm to select the hyper-parameter involved in the optimization.

\section{Related work}
\subsection{Existing packaging selection process}
\label{sec:currentapproach}
The decision to choose the package type for a product is currently based on a Keyword Based Approach (KBA), where the products are mapped to package types based on whether their title contains a predefined set of positive and negative keywords. Positive keywords work as enablers to ship products in inferior packaging types like polybags or NAP. Examples for positive keywords are helmet, diapers, mosquito net, bag pack, laptop sleeve, bedsheet, cushion, etc. with the assumption being that such products will have low in-transit damages due to inferior packaging. Negative keywords on the other hand prohibit opting for polybag. Examples for negative keywords are bone-china, detergent, harpic, protein supplements, etc. After manually analyzing the product titles, a suitable package type is identified. Another approach that is closely followed is the selection of package types using the historical data on damages. Here, the damage rates of products are collected on a monthly basis and based on set guardrails, the packaging rules are modified for high damaged products. In addition to being a slow, manual process, this is a reactive approach which does not work for many new products or products whose attributes have been modified recently. 

\subsection{Why not ordinal regression?}
\label{sec:or}
As the different packaging options can be graded in terms of their robustness, the package type forms an ordinal variable with implicit relative ordering between them. This observation naturally surfaces the following question: \emph{``Is predicting the optimal package type for a product just an instance of ordinal regression?''}. Though the answer appears to be a \emph{yes}, the problem lies in the lack of training data. The current assignment of package type to a product is known to be sub-optimal for most of the products w.r.t. trade off between shipping and damage costs. The ideal setting would demand that we have enough samples for every <product, package type> pair, so that one could assign true package type as the target label and perform ordinal regression on product features. This model could then be leveraged to predict package type for new products. Such an exercise would incur significant cost especially at the scale at which e-commerce company like Amazon operates and hence is practically infeasible. The lack of such ground truth data precludes us from performing ordinal regression analysis. We allude to this fact in Section~\ref{sec:pkgtyperecommendation}. \eat{Instead, we address the problem in a two-stage approach as elucidated below.}

\subsection{Comparison with standard machine learning approaches for package planning}
The work in~\cite{Knoll19} shows the adoption of machine learning (ML) in manufacturing industries for automated package planning. Given a training data with well defined labels of which package type has to be used for which product parts, the goal in these applications is to train a supervised ML model based on product characteristics, which are later used to predict package type for unseen products. Our current work differs from these approaches on the following factors:\\
(i) As described above, we do not have any ground truth data to learn a supervised ML model that directly predicts the package type given the product features. We have training label only at the shipment level that informs whether a product shipped in particular package type was damaged (1) or not (0).\\
(ii) There is natural ordering between the package types that should be enforced in any learning algorithm.\\
Given the above two constraints, we are not aware of any learning based framework that automatically chooses the best package type in linear time and is scalable to millions of products.

A majority of the work in logistics is around space optimization, which is broadly related to bin \emph{packing} algorithms \cite{Martinez15}, \cite{smallboxesbigdata}, not to be confused with the \emph{packaging} type selection problem. The aim of the former is to identify those set of products, each of a specific volume, that should be loaded together in a container, in a specific orientation, so that the number of container used is minimum. The bin packing problem has no notion of choosing the best package type for each product. Our work also has very little connection with the box size optimization problem~\cite{wilson1965}, where the goal is to determine the best box sizes that should be used to ship the products, so that the total shipment volume across all product is minimum. We do not optimize for the different sizes of the packages in the current work, but rather determine which package type is best suited for a product. Likewise, we do not forecast packaging demand like the methods developed in \cite{packagingforecasting}.

\section{Two-stage approach for optimal package selection}
\subsection{Stage 1: Estimating the transit damage probability of a product given a package type}
\label{sec:damageProb}
In this stage, we build a model to solve the following problem, \emph{``Given a product and a package type, what is the probability that a shipment of the product with that package type is likely incur costs due to damages?''}. These damage probabilities are computed for every <product, package type> pair as a product may never have been shipped using a particular (say hitherto unknown optimal) package type to directly retrieve it from the shipment data. In short, our model predicts $p(d | i, j)$ where $i$ refers the product, $j$ refers to the package type and $d$ is a variable indicating damage in transit, with $d=0$ denoting no damage and $d=1$ specifying a damage in transit.

For modeling, we considered historical shipment data where for every shipment we have a binary flag a.k.a. the target label indicating whether the shipment resulted in package related damage. We built this model using various metadata associated with the product as predictor variables. The following enumerate a sample set of attributes: product title, category, subcategory, product dimensions, weight, hazardous flag (indicating if product pertains to hazardous materials), fragile flag (denoting whether the product is fragile), liquid flag (representing if the product contains liquids), \% air in shipment computed as the difference between the package volume and the product volume, etc. Based on the above set of features we trained a model to predict the probability that the shipment using the particular package type will incur a damage. 

\subsubsection{Maintaining ordinal relationship between different package types}
\label{sec:ordinalrelationship}
The notion of graded robustness between package types correlates with the cost of the packaging material where the cost of packaging goes up if we opt for a more robust package type and vice versa. As there exists an ordinal relationship among  the various package types, i.e. they can be ordered in terms of their associated robustness, we need to impart this notion to our model while estimating the damage probabilities. Let $m$ and $n$ be the number of products and package type respectively and without loss of generality let the package type $j_k$ be inferior to $j_{k+1}$ represented by the ordinal relationship $j_1 \leq j_2 \leq \cdots \leq j_n$. During modeling we need to ensure that $ p(d | i, j_{k+1}) \leq p(d | i, j_k)$ for all products $i$. In other words, the prediction function needs to be \emph{rank monotonic} \cite{Li06} where the rank denotes the robustness of the package type. Note that we require the predictions to satisfy the ranking relationship only between the different package types associated for a given product and not across two different products. We achieve rank monotonicity by two means: (a)Augmenting the training data, and (b)Proper representation of the package type feature and imposing lower bound constraints on the corresponding model coefficients.  

\noindent Firstly, we append the modeling data as follows: \\
(i) For every damaged shipment, we create additional shipments with the same product and other inferior (less robust) package types and consider them to be damaged as well. This is to incorporate the notion that if a shipment of a product gets damaged with a particular package type, it is likely to get damaged in package types which are inferior in terms of robustness.\\
(ii) Likewise, for every shipment without any packaging related damages, we artificially introduce more shipments with the same product and other superior (more robust) package types and consider them to be not damaged as well. This is to incorporate the notion that if a shipment of a product does not get damaged with a particular package type, it is unlikely to get damaged in superior package types.

Appending the data set has an added advantage of creating many more samples for the positive damaged class (label = $1$), as typically very few shipments, less than $0.6\%$, incur packaging related damages. This in turn reduces the model variance as even the positive class is well represented. Secondly, expressing  the damage probability values in terms of the sigmoid function, namely $p(d | i, j) = \frac{1}{1+\exp(-f(\bz_i,j))}$ where $\bz_i$ denote the rest of input features barring the package type, we represent $f(.)$ as $f(\bz_i,j) = g(\bz_i)+ \beta_j$. Here $\{\beta_j\}_{j=1}^n$ are the $n$ model coefficients corresponding to each package type. Ensuring rank monotonicity is equivalent to constraining $\beta_k \geq \beta_{k+1}$. Expressing $\beta_{k} = \beta_{k+1} + \epsilon_k$, we enforce that $\epsilon_k \geq 0, \forall k \in \{1,2,\ldots,n-1\}$. In the event that $g(.)$ is linear, i.e., $ g(\bz_i) = \bw^T \bz_i$ as the case with Logistic Regression classifier, then for each package type $j_k$, we append the feature vector $\bz_i$ to create $\tilde{\bz}_{ik} = [\bz_i, \bp_k]$ where $\bp_k = [\underbrace{0,0,\dots,0}_{k-1},\underbrace{1,\dots,1}_{n-k+1}]$, augment the model coefficient vector $\bw$ to $\tilde{\bw} = [\bw, \epsilon_1,\dots,\epsilon_{n-1}, \beta_n]$, and express $f(\bz_i,j_k) = \tilde{\bw} ^T \tilde{\bz}_{ik}$. The vector $\tilde{\bw}$ is determined as part of the model training process under the constraint that $\epsilon_k \geq 0, \forall k$.

\subsection{Stage 2: Identifying the optimal package type for each product}
Optimally assigning the packaging type for each product involves finding the right balance between adopting a robust packaging and incurring more material and transport costs, and settling for an inferior option with a higher probability of in-transit damages leading to increased damage costs. We formulate this trade-off as an optimization problem. Given a packaging type assignment $j$ for a product $i$, the packing material cost $m(i,j)$ and the transportation cost $s(i,j)$ are known and readily available. The quantity $s(i,j)$ is known as the \emph{bill weight} and is proportional to the package volume. The net shipping cost, $C_{ship}(i,j)=  m(i,j)+s(i,j)$. The total shipment cost, $T_{ship}$, computed over all the products equals: $T_{ship}= \sum\limits_{i} C_{ship}(i,j) * s_{vel} (i)$, where $s_{vel} (i)$ is the sales velocity ---number of units sold in a specified period--- of the product $i$. Further, if a product $i$ associated with the package type $j$ is damaged in transit, we incur a net damage cost $C_{damage}(i)$. This damaged cost depends only on the product and independent of the package type used in the shipment.  Using the in-transit damage probability $p(d|i,j)$ determined in stage 1 (Section~\ref{sec:damageProb}), we estimate the damage cost as: $T_{damage}=\sum_{i} p(d|i,j) * s_{vel}(i) * C_{damage}(i)$.

Let us denote the current package type assignment of product $i$ by $j^{cur}$. According to the current package type assignment, the total cost due to in-transit damages is: $T_{damage}^{cur}=\sum\limits_{i} p(d|i,j^{cur}) * s_{vel} (i) * C_{damage}(i)$.
The objective of the optimization is to determine the package types such that $T_{ship}$ is minimized and at the same time $T_{damage}$ is not largely different from $T_{damage}^{cur}$ i.e., $T_{damage} \leq \gamma * T_{damage}^{cur}$, where $\gamma \geq 0$ sets the allowable tolerance w.r.t. $T_{damage}^{cur}$ and is determined by business requirements. 
\subsubsection{Mathematical formulation}
\label{sec:IP}
Let the variable $x_{ij}$ indicating whether a product $i$ is to be shipped using the package type $j$, be the $<i,j>$ entry of the binary matrix $X$.
\eat{
\begin{equation*}
X = 
\begin{bmatrix}
x_{11} & x_{12} &\cdots&x_{1n}\\
\vdots&\vdots &\vdots&\vdots \\
x_{m1} & x_{m2} &\cdots&x_{mn}
\end{bmatrix}.
\end{equation*}
}
These variables have to satisfy the following constraints, namely: $x_{ij} \in \{0,1\}$, $\forall i,j$ and $\sum_{j} x_{ij}=1$, $\forall i$. The first constraint states that a product is either shipped in a particular type $(x_{ij}=1)$ or not  $(x_{ij}=0)$. The second constraint specify that a product should be shipped using one and only one package type. In additional to the aforesaid binary constraints, we also need to specify infeasible conditions that preclude certain products to be shipped via certain modes of packaging. For instance, liquid, fragile and hazardous products can neither be recommended polybags nor be shipped without any packaging if they are not currently shipped in these package options. We enforce these infeasibility constraints by creating a mask matrix $M$ where we set $M_{ij}=1$ if product $i$ cannot be shipped in package type $j$ and $M_{ij} = 0$ otherwise. By imposing the constraint $\sum\limits_{i,j}M_{ij} * x_{ij} =0$, the optimization algorithm will be coerced to set $x_{ij}=0$ whenever $M_{ij}=1$, thereby meeting our infeasibility requirements. Letting $S_{ij} = C_{ship}(i,j) * s_{vel} (i)$ to be the net shipment cost when product $i$ is sent in package $j$, $D_{ij} = p(d|i,j) * s_{vel}(i) * C_{damage}(i, j)$ as the net damage cost when the shipment experiences an in-transit damage due to the packaging, $T= \gamma *T_{damage}^{cur}$, our objective can be mathematically expressed as:
\begin{align}
\label{eq:I}
&\min_{X} \sum_{i,j} S_{ij} * x_{ij} \hspace{10pt} s.t.   \hspace{10pt} \sum_{i,j} D_{ij} * x_{ij}\leq T\\
&\mbox{where,} \hspace{5pt} x_{ij} \in \{0,1\}, \hspace{2pt} \sum_{j} x_{ij}=1, \hspace{1pt} \forall i \mbox{  and  } \sum_{i,j} M_{ij} * x_{ij} =0. \nonumber
\end{align}

Computing the optimal solution for $X$ based on the Integer Programming (IP) formulation in eq.(\ref{eq:I}) is computationally expensive as it is a known $NP$-complete problem \cite{Papadimitriou98}. The IP formulation is definitely not scalable and is of very limited use for our setting. Hence, we \emph{do not} compute the solution for $X$ by solving eq.(\ref{eq:I}). We present the IP objective with the only intent of mathematically formulating and motivating our optimization problem. The direct minimization of the shipping cost, while enforcing that overall damage cost does not exceed the constant $T$, makes the setting easier to understand. We abstain from solving for $X$ based on this IP objective. 

A closer look into the constraints on $X$ reveals that, the constraints are only intra-product, i.e., across different packaging options for a given product and there are no inter-product constraints. This insight enables us to derive an equivalent formulation for eq.(\ref{eq:I}) whose solution, as we demonstrate, can be obtained via a simple linear time algorithm in $O(mn)$. To this end, let $S(X) = \sum\limits_{i,j} S_{ij} * x_{ij}$, $D(X) = \sum\limits_{i,j} D_{ij} * x_{ij}$, and consider the formulation:
\begin{align}
\label{eq:T}
\min_{X} \hspace{2pt} & E(X) = S(X) + \lambda D(X)\mbox{  s.t.,} \\
& x_{ij} \in \{0,1\}, \hspace{2pt} \sum_{j} x_{ij}=1, \hspace{1pt} \forall i \mbox{  and  } \sum_{i,j} M_{ij} * x_{ij} =0 \nonumber,
\end{align}
where the hyper-parameter $\lambda$ is a single globally specified constant \emph{independent} of the products and the package types. The constrained formulation in eq.(\ref{eq:I}) is known as the Ivanov formulation \cite{Ivanov76} and the objective in eq.(\ref{eq:T}) is referred to as the Tikhonov formulation \cite{Tikhonov77}. The equivalences between the two are specifically known for Support Vector Machines \cite{Oneto16}, \cite{Cortes95} where the optimization variable, the weight vector $\bw$, is continuous and is based on the Lagrange dual formulation. This approach does not work in our discrete setting where $X$ is binary valued. We need to establish this equivalence without invoking the Lagrange formulation and hence our proof methodology is substantially different.

Let $X_{\lambda}$ and $X_T$ be the optimal solutions for the hyper-parameters $\lambda$ and $T$ in Tikhonov and Ivanov formulations respectively. We now prove that under mild conditions on the shipment and damage cost values discussed in the Appendix, these two formulations are equivalent in the sense that for every $T$ in Ivanov, $\exists$ a value of $\lambda$ in Tikhonov such that both the formulations have the exact same optimal solution in $X$. To this end, we have the following lemmas:
\begin{lemma}
\label{lemma:E}
The value of the objective function $E\left(X_{\lambda}\right)$ at the optimal solution $\Xl$ strictly increases with $\lambda$.
\end{lemma}
\begin{lemma}
\label{lemma:D}
The overall damage cost $D\left(\Xl\right)$ \emph{[shipment cost $S\left(\Xl\right)$]} at the optimal solution $\Xl$ is a non-increasing \emph{[non-decreasing]} function of $\lambda$, i.e., if $\lambda_1 \leq \lambda_2$ then $D\left(X_{\lambda_1}\right) \geq D\left(X_{\lambda_2}\right)$ \emph{[$S\left(X_{\lambda_1}\right) \leq S\left(X_{\lambda_2}\right)$]}. Further, if $X_{\lambda_1} \neq X_{\lambda_2}$ then we get the strict inequality, namely $D\left(X_{\lambda_1}\right) > D\left(X_{\lambda_2}\right)$ \emph{[$S\left(X_{\lambda_1}\right) < S\left(X_{\lambda_2}\right)$]}.
\end{lemma}
Lemma~\ref{lemma:D} states that $D\left(\Xl\right)$ is a piece-wise constant function of $\lambda$ whose value decreases when the optimal solution changes. The length of the constant portion equals the range of $\lambda$ having the same optimal solution. Further, $D\left(\Xl\right)$ is discontinuous and points of discontinuity occurs at those values of $\lambda$ for which there are two different optimal solutions in $\Xl$. Please see the Appendix for details and the proofs. We establish the equivalence through the following theorems.
\begin{theorem}
\label{thm:T2I}
For every $\lambda$ in eq.(\ref{eq:T}), $\exists$ $T$ $(\gamma)$ in eq.(\ref{eq:I}) such that $X_{\lambda} = X_{T}$.
\end{theorem}
We define a quantity $\Delta$ to equal the largest change between the two values of $D\left(\Xl\right)$ at the points of discontinuity.\eat{It can be precisely defined as follows. For any $t > 0$, let $\Delta_t = \sup\limits_{r \in \mathbb{Z}^+} \left[D\left(X_{rt}\right)- D\left(X_{(r+1)t}\right)\right]$ where the values of $\lambda = rt$ jumps at intervals of $t$. Set $\Delta = \lim\limits_{t \rightarrow 0^+} \Delta_t$. Observing that $\Delta_t > 0$, non-increasing and bounded below by $0$, $\Delta$ is well-defined in the limit by the Monotone Convergence Theorem.} For our specific $D$ matrix, $\Delta \leq \max\limits_i \left[\max\limits_{j, M_{ij}=0} D_{ij} - \min\limits_{j, M_{ij}=0} D_{ij}\right]$. Armed with this definition, we now prove a mildly weaker equivalence in the opposite direction.
\begin{theorem}
\label{thm:I2T}
For every $T=\gamma*T_{damage}^{cur}$ in eq.(\ref{eq:I}) for which the optimal solution $X_T$ exists, one can find a $T^{\ast} \in \left[T, T+\Delta \right)$ such that for this value of $T^{\ast}$, $\exists \lambda$ in eq.(\ref{eq:T}) satisfying $X_{\lambda} = X_{T^{\ast}}$.
\end{theorem}

\subsubsection{Linear time algorithm}
\label{sec:linearTimeAlgo}
The primary advantage of this equivalence is that the Tikhonov formulation in eq.(\ref{eq:T}) enjoys a linear time algorithm compared to the Ivanov problem in eq.(\ref{eq:I}) which is $NP$-complete. To see this, note that the constraints in the variables $x_{ij}$ are only across the different package types $j$ given a product $i$ and there are no interaction terms between any two different products. Hence the optimization problem can be decoupled between the products and reduced to finding the optimal solution \emph{independently} for each product agnostic to others. For each product $i$, define the vector $\bx_i = [x_{i1}, x_{i2},\dots,x_{in}]$ and consider the optimization problem:
\begin{align}
\label{eq:productlevel}
\min_{\bx_i} \hspace{2pt} &\sum_j [S_{ij}+\lambda D_{ij}]x_{ij}\mbox{  s.t.,} \\
&x_{ij} \in \{0,1\}, \hspace{2pt} \sum_{j} x_{ij}=1\mbox{ and  } \sum_{j} M_{ij} * x_{ij} =0. \nonumber
\end{align}
Among all the package types where $M_{ij} = 0$, the minimum occurs at that value of $j = j^{\ast}_{i}$ where the quantity $S_{ij^{\ast}_{i}}+\lambda D_{ij^{\ast}_i}$ takes the least value. In other words, define $j^{\ast}_{i} = \argmin\limits_{j, M_{ij}=0} [S_{ij}+\lambda D_{ij}]$. Then $x_{ij^{\ast}_i} = 1$ and $x_{ik} = 0$, $\forall k \neq j^{\ast}_{i}$ is the optimal solution. As it involves a search over the $n$ values, its time complexity is $O(n)$ for each product. \emph{Hence the optimal solution $X_{\lambda}$ across all the $m$ products can be determined in $O(mn)$}. \eat{giving us the following corollary.
\begin{corollary}
For all values of $\lambda \geq 0$, the solution $X_{\lambda}$ for the Tikhonov formulation in eq.(\ref{eq:T}) can be computed in $O(mn)$.
\end{corollary} 
}
\subsubsection{Selection of the hyper-parameter $\lambda$}
It is often easier to specify a bound on the overall damage cost $D(\Xl)$ via the tolerance constraint $T = \gamma * T_{damage}^{cur}$ in the Ivanov formulation in eq.(\ref{eq:I}), as it is driven by business requirements such as customer satisfaction, impact of damages on downstream purchase behavior, etc. However, knowledge of $\gamma$ alone is of little value as the Ivanov formulation being $NP$-complete, is computationally expensive to solve and we rightly refrain from doing so. Instead, we determine the corresponding $\lambda$ through an efficient algorithm and then solve the Tikhonov formulation in eq.(\ref{eq:T}) in linear time as explained in Section~\ref{sec:linearTimeAlgo}.
Although no closed form expression exists relating the two, the non-increasing characteristic of $D\left(\Xl\right)$ in Lemma~\ref{lemma:D} can be leveraged to design a binary search algorithm for $\lambda$, as described in Algo.~\ref{algo:lambdaSearch}. The crux of our method is to repeatedly bisect the interval for the search space of $\lambda$ and then choose the subinterval containing the $\lambda$\eat{using the property of $D\left(\Xl\right)$}. The technique is very similar to the bisection method used to find the roots of continuous functions~\cite{bisection}. The user input $\rho$ in Algo.~\ref{algo:lambdaSearch} is the stopping criteria on the minimum required change in $\lambda$ values between successive iterations for the while loop to be executed. The number of iterations is inversely proportional to the magnitude of $\rho$. \eat{Smaller the value, more the number of binary searches and vice-versa.}

\begin{algorithm}[t]%[htpb]
 \caption{Algorithm to determine $\lambda$ given $T$ and stopping criteria $\rho$}
 \label{algo:lambdaSearch}
\begin{algorithmic}
\Function{DetermineLambda}{$T$, $\rho$}
\State \textbf{Set:} $\lambda_{min}=0$, $\lambda_{max} = \mbox{chosen high value}$, $\lambda_{mid} = \frac{\lambda_{min} + \lambda_{max}}{2}$, stoppingCriteria = False
\Do
	\State \textbf{Set:} $\lambda = \lambda_{mid}$
	\State \textbf{Determine:} Optimal solution $X_{\lambda}$ using the linear time algorithm.
	\If {$D\left(X_{\lambda}\right) < T$}
		\State $\lambda_{max} = \lambda_{mid}$
	\Else
		\State $\lambda_{min} = \lambda_{mid}$
       \EndIf
       \State \textbf{Recompute:} $\lambda_{mid} = \frac{\lambda_{min} + \lambda_{max}}{2}$
       \If {$\left(|\lambda_{mid} - \lambda| \leq \rho\right)$ or $(D\left(X_{\lambda}\right)==T)$}
		\State \textbf{Set:} stoppingCriteria = True 
	\EndIf      	
\doWhile {(stoppingCriteria==False)}
\State
\Return $\lambda$
\EndFunction
\end{algorithmic}
\end{algorithm}

\subsubsection{Package prediction for new products}
The definition of the net shipping and damage cost matrices includes the sales velocity term $s_{vel}(i)$, as the total shipment and damage costs across all products explicitly depend on the individual quantities of products sold. Hence the optimization problem in eq.(\ref{eq:I}) deliberately makes use of the sales velocity term folded into the $S_{ij}$ and $D_{ij}$ matrix entries. However, for new products, the sales velocity is unknown and needs to be forecasted; which is generally very difficult and at most times noisy \cite{Machuca14}. The lack of this term seems to preclude the new products from being part of the optimization in eq.(\ref{eq:I}). However, the equivalent Tikhonov formulation in eq.(\ref{eq:T}) comes to our rescue. Closely looking into the product-wise optimization problem in eq.(\ref{eq:productlevel}), note that $s_{vel}(i)$ appears in the same form (linearly) in both the $S_{ij}$ and $D_{ij}$ quantities and also \emph{does not} depend on the package type $j$. Hence it can be factored out and dropped from the optimization altogether. The equivalent formulation has revealed a key insight that once $\lambda$ is appropriately chosen, the optimal solution is \emph{independent} of the sales velocity. Hence for all new products $l$, we only need to compute quantities $\{S_{lj}, D_{lj}\}_{j=1}^n$ without factoring in $s_{vel}(i)$ and choose that package type $j^{\ast}_{l}$ with the least value of $S_{lj^{\ast}_{l}}+\lambda D_{lj^{\ast}_l}$ among the package types where $M_{lj} = 0$.

\section{Experimental results}
Our training data for stage 1 consists of shipments during a 3 month period in 2019. We augmented the data with artificially induced inferior and superior packaging types and their corresponding 1 and 0 target values. We opted for the Logistic Regression classifier to learn and predict the damage probabilities $p(d|i, j)$, as it enables us to interpret and explain the predictions. Importantly, its linearity (post the link function) endows the model with the notion of ordinal relationship between packages by appending the sample features $\bz_i$ with the package related features $\bp_k$ as elaborated in Section~\ref{sec:ordinalrelationship}. Though our classifier is linear in the feature space, we introduced non-linearity through polynomial transformations of the input features and having interactions between the product and the package features to create new (non-linear) features. In more than $100$ \emph{million} augmented training shipments, only $0.7\%$ shipments belonging to class $1$ incurred package related damages. We counter this huge class imbalance by specifying class specific weight values of $1-\tau$ and $\tau$ to classes $1$ and $0$ in the cross-entropy loss function where we set $\tau = 0.007$. We assessed the performance of our model on a test data consisting of $8$ \emph{million} shipments for about $600,000$ products, out of which only $0.6\%$ shipments incurred packaging related damages. The shipments in the test data occurred in a different time period w.r.t. training data. After augmenting the test data with artificially induced inferior and superior packaging types, the models performance on the area under the curve (AUC) metric was $\mathbf{0.902}$. 

\subsection{Calibration}
\begin{figure}
\begin{center}
\includegraphics[width = 0.9\linewidth]{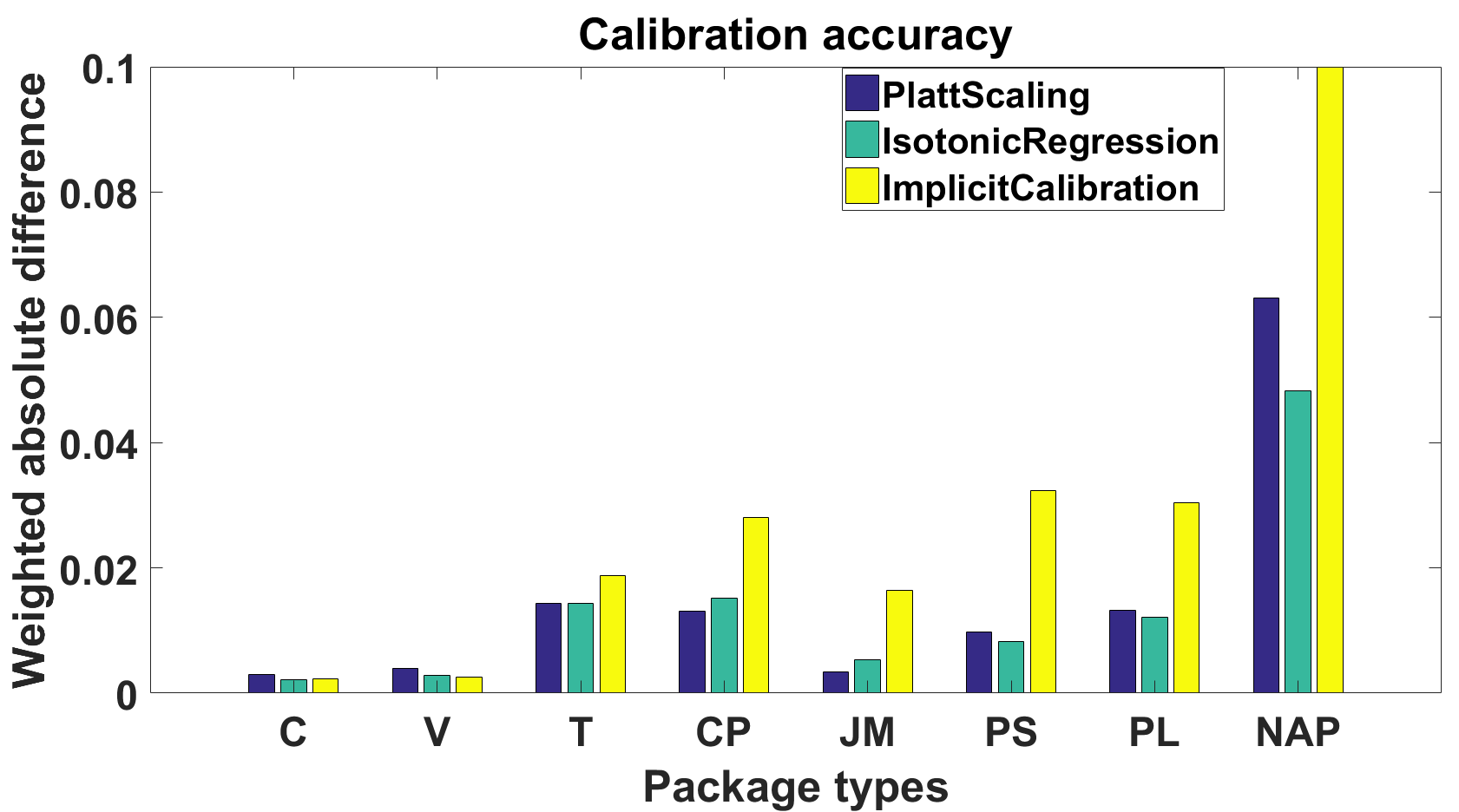}
\end{center}
\vspace{-0.2in}
\caption{Weighted absolute difference between estimated and true damage rates}
\label{table:calibration}
\end{figure}

\eat{Though we have quoted the AUC number above, it is a useful metric only for classification models where the purpose is to classify whether a shipment  (<product, package type>) will incur damages during transit.}Since we are interested in estimating the actual probability of damage rather than binary classification, the estimated raw damage probabilities  $p(d|i, j)$ need to be calibrated to reflect the true damage probabilities in the shipment data. This is more so, as we introduced class specific weights during training.\eat{Calibration involves learning the transformation function which when applied on the raw damage probabilities, transforms them closer to the ground truth. Using the raw damage probabilities computed on the test data and true $0/1$ labels of the shipments during the test period,} We used Isotonic Regression \cite{IsotonicRegression} to learn the calibration function. It yielded the smallest average \emph{log-loss} (log-loss = 0.0347)\eat{, as seen from Fig.~\ref{table:calibration}}, compared to the implicit calibration via the closed form expression derived in \cite{King01} (eq.(28)) for binary Logistic Regression models (log-loss = 0.0379) and Platt Scaling~\cite{PlattScaling} (log-loss = 0.0349). The log-loss for each shipment equals:$ - y \log \left(p_{cal}\right) - (1-y) \log \left(1-p_{cal}\right)$, where $y$ is the actual label and $p_{cal}$ is the calibrated damage probability. All these calibration methods significantly reduces the uncalibrated average log loss of $0.4763$. To assess the correctness of post-calibrated values, for each package type we bucketed its shipments into 20 quantiles based on their calibrated values. For each quantile, we computed the absolute difference between the actual damage rate and the average of the calibrated values, weighted these absolute differences proportional to the number of shipments in each quantile, and then summed them. Fig.~\ref{table:calibration} shows the summed, weighted absolute differences for each package type, for different calibration methods. Observe that Isotonic Regression has the lowest values across multiple package types. \emph{Such low difference values highlight the estimation accuracy of our post-calibrated damage probability values}.

\subsection{Package type recommendation}
\label{sec:pkgtyperecommendation}
For a dataset of about $250,000$ products in more than $10$ categories with active purchase history in Amazon, we determined their raw damage probability for all possible package type options and then calibrated them using Isotonic Regression. Table~\ref{table:avgdamageprob} shows the relative average damage probabilities computed across the products for each package type. The damage probability for shipment without packaging (NAP) is set to 1 and the values for other package types are scaled relatively. The business sensitive nature of these damage probabilities precludes us from disclosing their actual estimated values. Observe that our model has indeed learned the implicit ordering between the package types, where the superior package types like C and V have the lowest values and inferior package types like PS and PL have the highest. The predicted damage probabilities are then fed into our optimization algorithm that proposes optimal packaging type for all the products.

For each <product, package type> tuple, we identified the smallest size of that package type that could fit the product snugly. This reduces the shipment volume and also the shipment cost. Recall that by setting entries $M_{ij} = 1$ in the mask matrix $M$, we  can prevent the optimization from choosing the package type $j$ for product $i$. We set $M_{ij}=1$ for the following cases based on business rules: (i) products which due to its large size and volume cannot be shipped even in the largest container of certain package types, equivalent to setting the corresponding $S_{ij} = \infty$, (ii) liquid products from being shipped in JM, PS, PL or NAP; restricting fragile products from being sent in T, CP, JM, PS, PL or NAP; disallowing hazardous products to be shipped in PS, PL or NAP if these products are not currently shipped in these package types (the latter condition is required as these flags can sometimes be erroneously set), (iii) inferior package types compared to the current selection i.e., $j < j^{cur}$ for products (with active purchase history) having high damages in the current package type, (iv) superior package types $j > j^{cur}$ if $S_{ij} > S_{ij^{cur}}$ for products with very low damages in their current packaging type, (v) sensitive products belonging to certain categories from being sent in NAP without any packaging etc.

\begin{figure}
\begin{center}
\includegraphics[width = 0.95\linewidth]{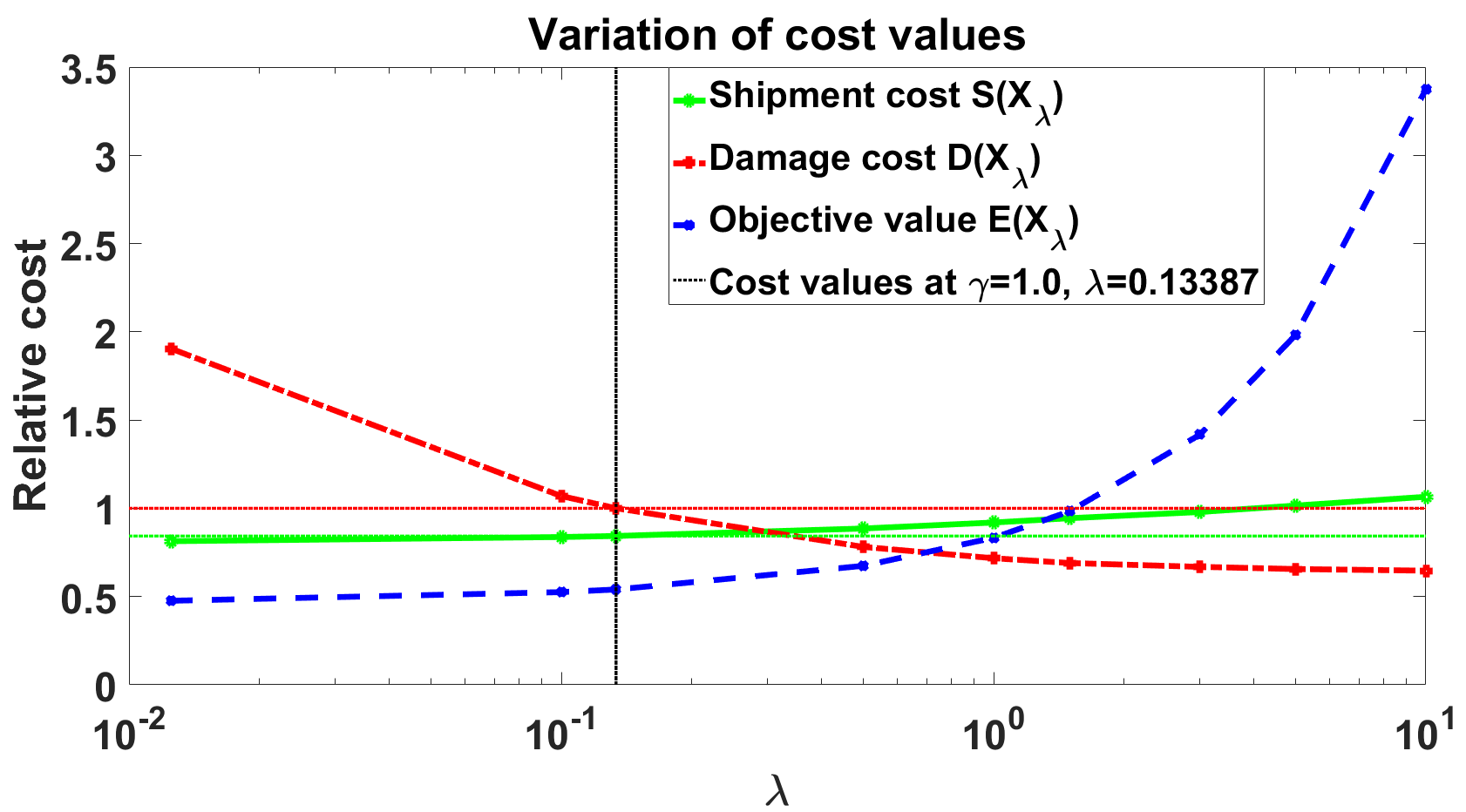}
\end{center}
\vspace{-0.2in}
\caption{Variation of relative cost values with $\lambda$}
\label{table:costWithLambda}
\end{figure}

To corroborate our theoretical results in Lemmas~\ref{lemma:E} and \ref{lemma:D}, we ran the Tikhonov formulation in eq.(\ref{eq:T}) for different values of $\lambda$, each in linear-time, and plot the results in Fig.~\ref{table:costWithLambda}. The values of $S(\Xl)$ and $D(\Xl)$ are scaled relative to the total shipment and damage costs from using the current package type, respectively. A value greater (lesser) than $1$ indicates that these costs will be higher (lower) compared to the current levels when the products are shipped based on the model recommended package types. Similarly, $E(\Xl)$ is scaled relative to the sum of current shipment and damage costs. Observe that the trends of $E(\Xl)$, $D(\Xl)$ and $S(\Xl)$ as we increase $\lambda$ are in accordance with the claims made in Lemmas~\ref{lemma:E} and \ref{lemma:D}. To verify the equivalence relations between the Ivanov and Tikhovov formulations stated in Theorems~\ref{thm:T2I} and \ref{thm:I2T}, we implemented the Integer Programming for Ivanov by setting $\gamma = 1.0$ using the CVXPY package \cite{cvxpy}. We then determined the value of corresponding $\lambda$ by executing our binary search method (Algo.~\ref{algo:lambdaSearch}) for $\rho=0.001$ and $\lambda_{max}=1000$. The algorithm met the stopping criteria in $19$ iterations and returned with $\lambda = 0.13387$. The identical results for $(\lambda=0.13387, \gamma=1.0)$ in columns VII and VIII of Table~\ref{table:avgdamageprob} is a testimony to this equivalence relationship. We validated this equivalence for other values of $\gamma$ using our binary search algorithm and obtained similar results. In Fig.~\ref{table:costWithLambda}, note that though the net damage cost $D(\Xl)$ for $\lambda = 0.13387$ (and for $\gamma=1.0$) exactly matches the cost value computed from using the current package types (ratio = 1 marked in horizontal red dotted line), the shipping cost $S(\Xl)$ is smaller than the current shipment cost (ratio = 0.843 marked in horizontal green dotted line). In other words, we are able to reduce the shipping cost from the current value without further increasing the damage cost. This again points to the fact that the existing product to package type mappings are sub-optimal, preventing us from pursuing the path of ordinal regression as explained in Section~\ref{sec:or}.

For each package type in Table~\ref{table:avgdamageprob}, we show the ratio of number of products mapped to that package type by our algorithm and the number of products currently assigned to the package type, for different $\lambda$ values. For instance if 100 products are currently shipped in package type C and our model recommends using C for 120 products, the ratio will equal $1.2$. A number greater (lesser) than $1$ denotes higher (lesser) recommendation of that package type compared to the current usage. Note that as we increase $\lambda$ giving more importance to damage cost, the ratio for superior package types such as C and V steadily increases, and this trend is reversed for inferior packaging options such as PL and NAP. This shift is as expected since the damage rate and the damage cost decrease at higher $\lambda$ values. In Table~\ref{table:productcategories} we show the ratio between the number of products recommended to be sent in a particular package type computed at $\lambda=1.5$ and the number of products currently shipped in these package types for different product categories. The value $0/0$ means no product of that category is currently shipped in the specific package type and our model does not recommend it either. The true counts \eat{of the number of package types used for each product category }are confidential and cannot be disclosed. Observe that for liquid, fragile and hazardous products, the ratio is less than $1$ for inferior package types such as JM, PS, PL and NAP, indicating that our method recommends lesser usage of these options for these kinds of products. Many electronics products in column VI with high damage probability are moved to the most superior $C$ package type, further contributing to the decreased damage rate of $24\%$ as observed in Section~\ref{sec:actualimpact}.
\begin{table*}
\centering
\caption{Relative avg. calibrated damage probabilities and change in product mappings for package types}
\label{table:avgdamageprob}
\begin{tabular}{|c|c|c|c|c|c|c|c|}
\hline
I&II&III&IV&V&VI&VII&VIII\\
\hline
&\emph{Package type} & \emph{Relative avg.}  &\emph{Ratio} &\emph{Ratio}& \emph{Ratio}& \emph{Ratio}&\emph{Ratio}\\
&& \emph{damage} & \emph{for} &\emph{for} & \emph{for} &\emph{for}& \emph{for} \\
&&\emph{probability}&$\lambda=0.5$&$\lambda=1$&$\lambda=1.5$&$\lambda=0.13387$&$\gamma=1$\\
\hline
Superior&Carton box(C) & 0.022&0.915&1.176&1.314&0.499&0.499\\
\cline{2-8}
package type&Variable height(V) & 0.027&0.566&0.676&0.740&0.392&0.392\\
\cline{2-8}
\multirow{4}{*}{$\downarrow$}&T-folder(T) & 0.043&1.227&1.282&1.298&0.898&0.898\\
\cline{2-8}
&Custom pack(CP) & 0.112&2.420&2.438&2.418&2.174&2.174\\
\cline{2-8}
&Jiffy mailer(JM) & 0.174&0.408&0.503&0.539&0.236&0.236\\
\cline{2-8}
&Small polybag(PS) & 0.447&1.586&1.360&1.238&1.955&1.955\\
\cline{2-8}
Inferior&Special polybag(PL) & 0.448&1.043&0.906&0.798&0.973&0.973\\
\cline{2-8}
package type&No packaging(NAP) & 1.0&1.748&1.144&0.940&3.635&3.635\\
\hline
%\vspace{-0.1in}
\end{tabular}
\end{table*}
\subsection{Impact analysis from actual shipment data}
\label{sec:actualimpact}
The numbers quoted below are excerpts from the actual shipment data, where for $130,000$ products contributing to 21\% shipments, their current package type was changed to the model's recommendation. We used the proposed package type obtained for $\lambda=1.5$ ($\gamma=0.69$), thus giving higher weight to reducing damage costs. The rationale being that receiving damage products negatively affects the customer trust in e-commerce companies and could affect their downstream purchase behavior. \eat{Another interpretation is that, we are assuming that the average \emph{causal} impact of receiving a damaged product because of poor packaging on customer's future purchase behavior, is approximately $0.5$ times the damage cost, and hence the net damage effect is 1.5 times the damage cost. When stated in the framework of Rubin Causal model~\cite{Causal}, the treatment is that of receiving a damaged product due to insufficient packaging, the potential outcomes are customer's future spends and by setting $\lambda = 1.5$ we \emph{implicitly assume} that the \emph{Average Treatment Effect} $=0.5 \times$ $damage$  $cost$. Estimating this casual impact is beyond the scope of this current work and we plan to investigate it in the future.} 
When these shipments were compared against those where the original package type was used, we observed the following significant positive impacts: (i) Decrease in damage rate by $\mathbf{24\%}$, (ii) Decrease in transportation cost per shipment by $5 \%$, (iii) Salability of products undelivered to customer because of transit damages improved by $3.5\%$. The only negative impact was that the material cost of the shipping supplies increased  by $2\%$, as many products were moved to superior package types to reduce damages. \eat{(by $\mathbf{24\%}$ as noted above)}
\begin{table*}[ht!]
\vspace{-0.1in}
\centering
\caption{Relative change in product mappings across different categories}
\label{table:productcategories}
\begin{tabular}{|c|c|c|c|c|c|c|c|}
\hline
I&II&III&IV&V&VI&VII&VIII\\
\hline
&\emph{Package type} & \emph{Liquid}&\emph{Fragile} &\emph{Hazardous} &\emph{Electronics}& \emph{Kitchen}&\emph{Beauty}\\
&&\emph{products}&\emph{products}&\emph{products}&\emph{category}&\emph{category}&\emph{category}\\
\hline
Superior&Carton box(C) &0.929&1.794&1.975&9.525&1.214&0.927\\
\cline{2-8}
package type&Variable height(V) & 0.754&1.047&0.915&1.017&0.859&0.478\\
\cline{2-8}
\multirow{4}{*}{$\downarrow$}&T-folder(T) &1.496&1.286&1.281&2.302&1.427&1.221\\
\cline{2-8}
&Custom pack(CP) &0/0&1.383&1.081&1.873&2.855&0.818\\
\cline{2-8}
&Jiffy mailer(JM) &0.519&0.486&0.488&0.392&0.695&0.928\\
\cline{2-8}
&Small polybag(PS) &0.000&0.821&0.843&1.203&1.263&1.720\\
\cline{2-8}
Inferior&Special polybag(PL) &0/0&0.541&0.500&0.863&1.124&4.000\\
\cline{2-8}
package type&No packaging(NAP) &0/0&0.598&0.806&0.737&0.701&2.000\\
\hline
\end{tabular}
\end{table*}
\section{Conclusion and future work}
We presented a two-stage approach to recommend optimal packaging type for products, where we first estimated the calibrated damage probabilities for every <product, package type> tuple and then fed them into our linear-time optimization algorithm to select the best type. The binary search algorithm efficiently computes the trade-off parameter $\lambda$ given the value $\gamma$ in the Ivanov formulation.

In many scenarios, the extent of damages depend on the distance shipped, the air/ground mode of transportation used, the quality of roads along the route, the handling by the courier partners, the location of the warehouses or even the time of year as during the monsoon season, more protection against water or moisture may be needed for some products. In addition, protective packaging could be recommended for specific customers who are highly valued or who had negative delivery experiences in the past. Going forward, we would like to lay emphasis on predicting the optimal packaging type based not only on the product, but using several aforementioned additional factors relating to a specific shipment of an item to a customer. Additionally, we would like to estimate the causal impact~\cite{Causal} of receiving damage products on customer's spend patterns and factor it into our optimization algorithm.

\bibliography{packaging}
\bibliographystyle{icml2020}

\appendix
\section*{Appendix}
\section{Non-collinearity conditions}
We make the following two non-collinearity conditions on the shipment and damage costs, in the entries of $S$ and $D$ matrices respectively, to establish the equivalence between the Ivanov and Tikhonov formulations stated in eqs.~(\ref{eq:I}) and (\ref{eq:T}). These conditions ensure that for every choice of $\lambda$, there are at most two possible solutions for $X_{\lambda}$ differing in packaging assignment \emph{on exactly one product}. These conditions are only a \emph{minor technicality} required to mathematically and precisely establish this equivalence. They do not play any role in the actual implementation and have no bearing on the quality of the results. We encourage the reader to look through the proofs detailed below to appraise their need. Further, as the entries in $S$ and $D$ matrices are arbitrary real numbers, these non-collinearity conditions are almost surely valid with probability one. So for all practical purposes they can be considered to be true.
\begin{assumption}
\label{ass:noncol1}
For every product $i$, no three points $(D_{ij}, S_{ij})$ across different package types when represented in the cost 2D plane are collinear. 
\end{assumption}
\begin{assumption}
\label{ass:noncol2}
For every $\lambda$, there does not exist two products $i_1$ and $i_2$ for which we can find corresponding package types $j_{i_11}$, $j_{i_12}$, $j_{i_21}$, and $j_{i_22}$ that \emph{simultaneously} satisfy the equations:
\begin{align}
S_{i_1j_{i_11}} + \lambda D_{i_1j_{i_11}} &= S_{i_1j_{i_12}} + \lambda D_{i_1j_{i_12}} \\
S_{i_2j_{i_21}} + \lambda D_{i_2j_{i_21}} &= S_{i_2j_{i_22}} + \lambda D_{i_2j_{i_22}}.
\end{align}
\end{assumption}
\section{Proof of Lemma~\ref{lemma:E}}
Consider two values $\lambda_1$ and $\lambda_2$ such that $\lambda_1 < \lambda_2$. Since $X_{\lambda_1}$ and $X_{\lambda_2}$ are the corresponding optimal solutions for these $\lambda$ values in eq.(\ref{eq:T}) we have
\begin{align*}
E\left(X_{\lambda_1}\right) &= S\left(X_{\lambda_1}\right) + \lambda_1 D\left(X_{\lambda_1}\right) \\
&\leq S\left(X_{\lambda_2}\right) + \lambda_1 D\left(X_{\lambda_2}\right) \\
&< S\left(X_{\lambda_2}\right) + \lambda_2 D\left(X_{\lambda_2}\right) = E\left(X_{\lambda_2}\right) 
\end{align*}
and the result follows.
\qed
\section{Proof of Lemma~\ref{lemma:D}}
We derive the proof for the damage cost $D(\Xl)$. The proof for $S(\Xl)$ follows along similar lines. Let $\Xlo$ and $\Xlt$ be the optimal solutions at the two values $\lambda_1 < \lambda_2$. We then arrive at the inequalities
\begin{align}
\label{eq:ineq1}
S\left(X_{\lambda_1}\right) + \lambda_1 D\left(X_{\lambda_1}\right) & \leq S\left(X_{\lambda_2}\right) + \lambda_1 D\left(X_{\lambda_2}\right), \\
\label{eq:ineq2}
S\left(X_{\lambda_2}\right) + \lambda_2 D\left(X_{\lambda_2}\right) & \leq S\left(X_{\lambda_1}\right) + \lambda_2 D\left(X_{\lambda_1}\right). 
\end{align}
On summing the two inequalities we have
\begin{align*}
&\lambda_1 D\left(X_{\lambda_1}\right) + \lambda_2 D\left(X_{\lambda_2}\right) \leq  \lambda_1 D\left(X_{\lambda_2}\right) +  \lambda_2 D\left(X_{\lambda_1}\right) \\
\implies &  \lambda_1 \left[D\left(X_{\lambda_1}\right) - D\left(X_{\lambda_2}\right) \right] \leq \lambda_2 \left[D\left(X_{\lambda_1}\right) - D\left(X_{\lambda_2}\right) \right].
\end{align*}
As $\lambda_1 < \lambda_2$, it follows that $D\left(X_{\lambda_1}\right) \geq D\left(X_{\lambda_2}\right)$ proving the non-increasing nature of $D(\Xl)$.

We establish the strict inequality via contradiction. Letting $D(\Xlo) = D(\Xlt)$ in the inequalities~\ref{eq:ineq1} and \ref{eq:ineq2}, we have $S\left(X_{\lambda_1}\right)  \leq S\left(X_{\lambda_2}\right)$ and $S\left(X_{\lambda_2}\right)  \leq S\left(X_{\lambda_1}\right)$, $\implies S\left(X_{\lambda_1}\right) = S\left(X_{\lambda_2}\right)$. Specifically we deduce that $\Xlo$ and $\Xlt$ are both solutions for $\lambda_1$ and $\lambda_2$. Let $j^1(i)$ and $j^2(i)$ be the assigned package types for product $i$ at $\lambda_1$ and $\lambda_2$ respectively. If $\Xlo \neq \Xlt$, then $\exists i$ such that $j^1(i) \neq j^2(i)$ and
\begin{align*}
S_{ij^1(i)} + \lambda_1 D_{ij^1(i)} &= S_{ij^2(i)} + \lambda_1 D_{ij^2(i)}, \\
S_{ij^1(i)} + \lambda_2 D_{ij^1(i)} &= S_{ij^2(i)} + \lambda_2 D_{ij^2(i)}.
\end{align*}
as these solutions can differ only for this specific product $i$ in accordance with the condition~\ref{ass:noncol2}. So $S_{ij^1(i)} = S_{ij^2(i)}$ and $D_{ij^1(i)} = D_{ij^2(i)}$. As points in the $2D$ cost plane, $\left(D_{ij^1(i)},S_{ij^1(i)}\right) = \left(D_{ij^2(i)},S_{ij^2(i)}\right)$ violating the non-collinearity condition~\ref{ass:noncol1}. So if $D(\Xlo) = D(\Xlt), \implies \Xlo =  \Xlt$ and the results follows.
\qed
\section{Proof of Theorem~\ref{thm:T2I}}
For the given $\lambda$, set $T=D(\Xl)$ as the tolerance value for the Ivanov formulation in eq.(\ref{eq:I}). At the solution $X_{T}$ for eq.(\ref{eq:I}), we observe that $D(X_T) \leq T$ as it should satisfy the tolerance constraint. Since the minimum occurs at $X_T$, we further have $S(X_T) \leq S(\Xl)$ and hence $S(X_T) + \lambda D(X_T) \leq S(\Xl) + \lambda D(\Xl)$. Since $\Xl$ is the solution for eq.(\ref{eq:T}), it follows that $S(X_T) + \lambda D(X_T) = S(\Xl) + \lambda D(\Xl)$ implying that $S(X_{T}) = S(\Xl)$ and $D(X_{T}) = D(\Xl)$. Hence both $\Xl$ and $\XT$ are solutions for both the Ivanov and Tikhonov formulations. If $\Xl \neq \XT$, then at this point of discontinuity, $D(\Xl) \neq D(\XT)$ in accordance with Lemma~\ref{lemma:D} resulting in a contradiction. Therefore $\Xl = \XT$ giving us the desired result.
\qed
\section{Proof of Theorem~\ref{thm:I2T}}
Given a tolerance value $T$ in eq.(\ref{eq:I}), let $X_T$ be the Ivanov solution. Recalling that $D(\Xl)$ is a non-increasing function of $\lambda$, define
\begin{align*}
\lambda_{inf} &\equiv \inf \{ \lambda \mbox{ s.t. } D(\Xl) \leq T), \mbox{ and}\\
\lambda_{sup}&\equiv \sup \{ \lambda \mbox{ s.t. } D(\Xl) \geq T)\}.
\end{align*}
We first show that $\lambda_{inf} = \lambda_{sup} = \tilde{\lambda}$ by considering two different cases. If $D\left(X_{\lambda_{inf}}\right) = D\left(X_{\lambda_{sup}}\right)$, then pursuant to Lemma~\ref{lemma:D} $X_{\lambda_{inf}} = X_{\lambda_{sup}}$ and hence $\lambda_{inf} = \lambda_{sup} = \tilde{\lambda}$ as these are defined to be points of extremities. When $D\left(X_{\lambda_{inf}}\right) \neq D\left(X_{\lambda_{sup}}\right)$, then it is the point of discontinuity where both $X_{\lambda_{inf}}$ and $X_{\lambda_{sup}}$ are two different solutions at the same value of $\lambda_{inf} = \lambda_{sup} = \tilde{\lambda}$.

If $D\left(X_{\lambda_{inf}}\right) = T$ or $D\left(X_{\lambda_{sup}}\right) = T$, then from Theorem~\ref{thm:T2I} we have (one of) $\Xld = X_{T}$ and the result follows. Otherwise, as $\Delta$ is defined to be the largest change between the consecutive values of $D(\Xl)$, we find $D\left(X_{\lambda_{sup}}\right) -D\left(X_{\lambda_{inf}}\right) \leq \Delta$. As $D\left(X_{\lambda_{inf}}\right) < T <  D\left(X_{\lambda_{sup}}\right)$ (note the strict inequality), we get 
\begin{align*}
D\left(X_{\lambda_{inf}}\right) &\in (T- \Delta, T), \mbox{ and }\\ 
D\left(X_{\lambda_{sup}}\right) &\in (T, T+\Delta).
\end{align*}
Ergo, in either case we can deduce that $\exists$ $T^{\ast} \in [T, T+\Delta)$ satisfying $D\left(\Xld \right) = T^{\ast}$ at $\Xld = X_{\lambda_{sup}}$. If $X_{T^{\ast}}$ is the Ivanov solution in eq.(\ref{eq:I}) for the tolerance value $T^{\ast}$, then from Theorem~\ref{thm:T2I} we have $\Xld = X_{T^{\ast}}$ and the result follows. 
\qed

\section{Behavior of the cost functions with $\lambda$}
\begin{figure}
\begin{center}
\begin{tabular}{c}
	\includegraphics[width=0.9\linewidth]{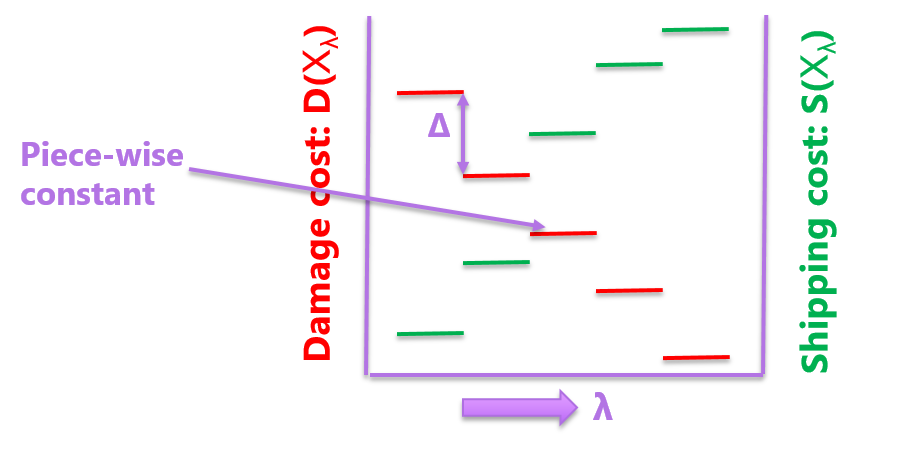}
\end{tabular}
\end{center}
\caption{Graph of damage and shipping costs at the solution $\Xl$ with $\lambda$.}
\label{fig:costfunctionwithlambda}
\end{figure}

As described above, both $D\left(\Xl \right)$ and $S\left(\Xl \right)$ are piece-wise constant functions of $\lambda$ whose values respectively decrease and increase when the optimal solution changes. The length of the constant portion equals the range of $\lambda$ having the same optimal solution. Further, both $D\left(\Xl \right)$ and $S\left(\Xl\right)$ are discontinuous functions and points of discontinuity occurs precisely at those values of $\lambda$ for which there are two different optimal solutions in $\Xl$. Figure~\ref{fig:costfunctionwithlambda} paints a visual description of this behavior.
\end{document}